\documentclass[letterpaper]{article} 
\usepackage[submission]{aaai25}  
\usepackage{times}  
\usepackage{helvet}  
\usepackage{courier}  
\usepackage[hyphens]{url}  
\usepackage{graphicx} 
\usepackage{natbib}  
\usepackage{caption} 
\usepackage{algorithm}
\usepackage{algorithmic}

\usepackage{amsmath}
\usepackage{xspace}
\usepackage{amssymb}
\usepackage{multirow}
\usepackage{enumitem}
\usepackage{makecell}
\usepackage{url}            
\usepackage{booktabs}       
\usepackage{amsfonts}       
\usepackage{nicefrac}       
\usepackage{microtype}      
\usepackage{xcolor}         

\usepackage{newfloat}
\usepackage{listings}
\DeclareCaptionStyle{ruled}{labelfont=normalfont,labelsep=colon,strut=off} 
\floatstyle{ruled}
\newfloat{listing}{tb}{lst}{}
\floatname{listing}{Listing}
\newcommand{\ie}{\textit{i.e.,}\xspace}
\newcommand{\eg}{\textit{e.g.,}\xspace}

\newcommand{\paratitle}[1]{\vspace{0.8ex}\noindent \textbf{#1}}

\newcommand{\llama}{Llama\xspace}

\title{FLM-101B: An Open LLM and How to Train It with \$100K Budget}

\author{
  Xiang Li\textsuperscript{1\textdagger}, 
  Yiqun Yao\textsuperscript{1\textdagger},
  Xin Jiang\textsuperscript{1\textdagger}, 
  Xuezhi Fang\textsuperscript{1\textdagger}, 
  Xuying Meng\textsuperscript{2},  \\
  \textbf{Siqi Fan\textsuperscript{3}, Peng Han\textsuperscript{3}, Jing Li\textsuperscript{4}, 
  Li Du\textsuperscript{1}, Bowen Qin\textsuperscript{1},
  Zheng Zhang\textsuperscript{1},} \\ 
  \textbf{Aixin Sun\textsuperscript{5}, Yequan Wang\textsuperscript{1$*$}}\\
  $^{1}$Beijing Academy of Artificial Intelligence, Beijing, China\\
  $^{2}$Institute of Computing Technology, Chinese Academy of Sciences, Beijing, China\\
  $^{3}$University of Electronic Science and Technology of China, Chengdu, China\\
  $^{4}$Harbin Institute of Technology, Shenzhen, China\\
  $^{5}$School of Computer Science and Engineering, Nanyang Technological University, Singapore
}

\begin{document}

\maketitle

\begin{abstract}
Large language models (LLMs) are considered important approaches towards foundational machine intelligence, achieving remarkable success in Natural Language Processing and multimodal tasks, among others. However, the carbon footprints and financial costs originating from heavy pre-training computation is a non-negligible issue. Progressive training methods, inspired by the neurogenesis process that grows neural structures, have shown potential to accelerate LLM pre-training. However, the algorithms, implementation, and practices for progressively training LLMs beyond 100B parameters remain underexplored. In this paper, we show that our model, namely FLM-101B, trained with our growth strategy under a budget of \$100K, reaches 80\% of the baselines' performances with only 10\% of their floating-point operations. We believe that further studies on progressive training will benefit the community by cutting down the costs and promoting green AI. The checkpoint of FLM-101B is publicly available.

\end{abstract}

\section{Introduction}
\label{sec:intro}

Large language models (LLMs) \cite{radford2018improving, llama,bert,t5} have consistently demonstrated their efficacy across a spectrum of applications, especially in language processing ~\cite{cort,cofenet,sig,dc-net} and multimodal tasks~\cite{DBLP:journals/corr/abs-2303-18223,netgpt}.
Despite variations in architectural designs, a universal challenge confronting all LLMs is the escalating cost associated with their training. Recent trends indicate a shift towards utilizing larger amounts of data (\eg 1.4T tokens for \llama-1~\cite{llama}, 2T tokens for \llama-2~\cite{llama-2}, and 15T tokens for \llama-3 \cite{llama3}). Meanwhile, the sizes of open-sourced models continue to increase \cite{falcon,deepseek,mistral-8-22}. Consequently, a major focus within LLM research is the development of innovative methodologies that effectively mitigate training expenses, aligning with the broader objectives of Green AI \cite{greenai}.

\begin{figure}
    \centering
    \includegraphics[scale=0.35]{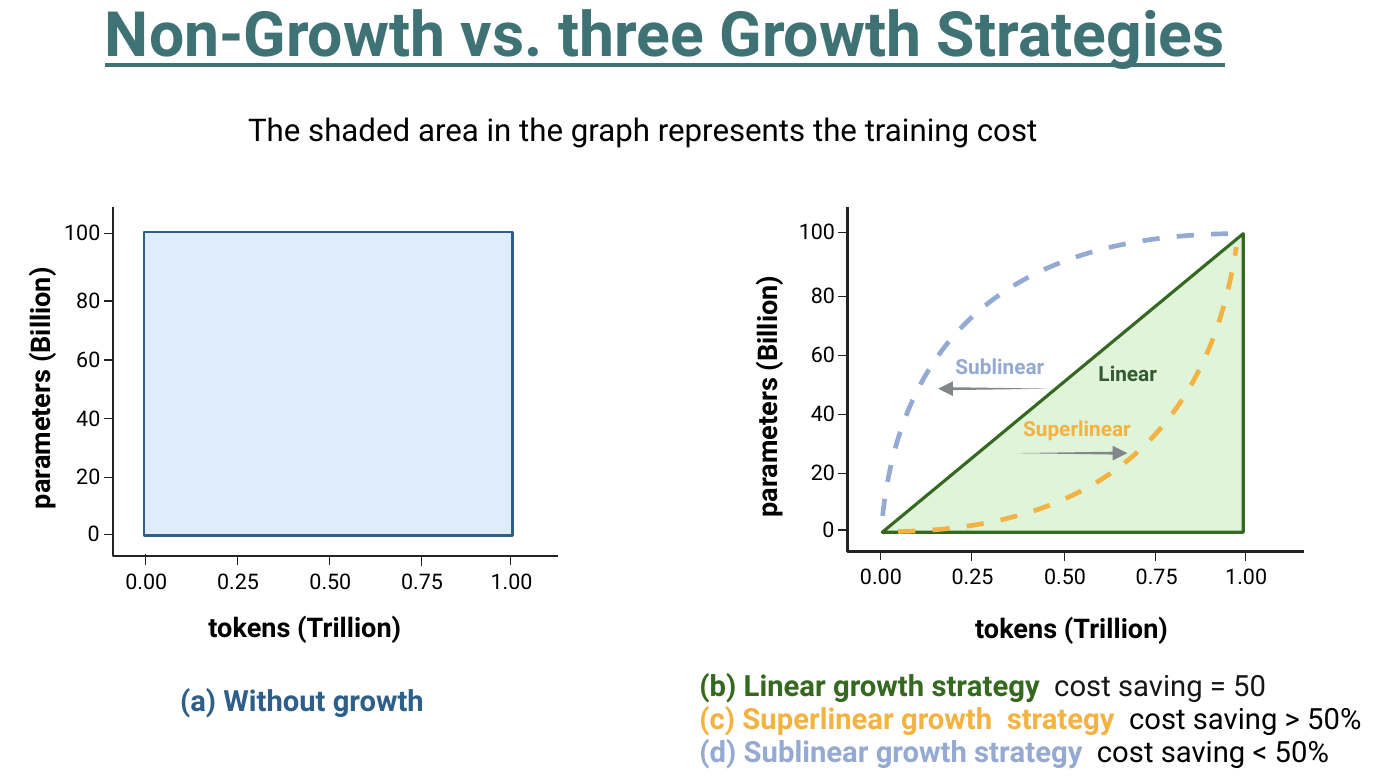}
    \caption{\textbf{An overview of different growth strategies.} (a): a baseline with constant number of parameters. (b): a straightforward linear growth strategy, cost-saving being exactly 50\%; (c): a superlinear strategy with $>$ 50\% cost saving; (d): sublinear strategy saving the cost by less than 50\%.}
    \label{fig:growth}
\end{figure}

In this paper, we present our exploration to train an LLM at the 100B-parameter scale using a \textit{growth strategy} inspired by previous research on progressive learning~\cite{stacking,compound,msg} and neurogenesis \cite{neurogenesis}. ``Growth'' means dynamic expansion of the parameter number count, from small to large, through the training progresses. Figure~\ref{fig:growth} illustrates three typical growth strategies: linear, sublinear, and superlinear. As the FLOPs of LLMs are approximately proportional to their number of parameters~\cite{chinchilla}, the area under the parameter curve represents the computational cost of training. 

While existing studies on scaling laws \cite{chinchilla} suggest that training a smaller model with more data may potentially result in higher scores on some tasks under a fixed FLOPs budget, they mainly consider the scenarios where model sizes are fixed through training. We believe that verifying the feasibility of a growth strategy \cite{compound, staged, bert2bert, msg} for extremely large models would be an important completion to scaling laws. To maximize computational efficiency, we strategically focus on implementing an aggressive growth strategy (Figure~\ref{fig:growth}~(c)) for sanity check. We adapt the MSG~\cite{msg} growth operators to train a model at 100B+ scale. We fix our budget to be \$100K with 192 A800 GPUs.

At the 100B scale, it is impractical to conduct strict head-to-head comparison with the same model trained with fixed size from scratch. Instead, we compare our grown model, namely FLM-101B, with the existing first-generation 100B+ language models \cite{gpt3, glm-130b}. Our model is trained with around 300 billion English and Chinese tokens, aligning it with these predecessors in terms of data scale. We first evaluate on the knowledge-oriented benchmarks~(\ie MMLU~\cite{mmlu} and C-Eval~\cite{c-eval}). Nevertheless, such evaluation may not comprehensively reflect the models' capability: it is difficult to distinguish whether the models recall a piece of knowledge or possess the capacity for reasoning and/or inference. Borrowing some ideas from Intelligence Quotient~(IQ) tests (\ie Perceptual Reasoning and Working Memory \cite{watkins}), we consolidate another range of evaluation, including \textit{symbolic mapping}~\cite{DBLP:journals/corr/abs-2305-08298}, \textit{rule understanding}, \textit{pattern mining}, and \textit{anti-interference} evaluations. 

We believe these tasks are less likely to be affected by data leakage or memorization, offering a more nuanced insight into the model's cognitive abilities beyond mere knowledge retrieval.

To summarize, the paper has made the following contributions. First, to the best of our knowledge, this is the first attempt to use a \textit{growth strategy} to train an LLM with 100B+ parameters \textbf{from scratch}. The training costs only \underline{100,000} US dollars. Second, we demonstrate details for addressing the instability issues via improving training objectives, hyperparameter search, and function-preserving growth. Third, we conduct extensive evaluations, including both the commonly used knowledge-oriented benchmarks and the new range of evaluations inspired by IQ tests. Experimental results show that, despite its low training cost, FLM-101B is competitive and robust. Lastly, we will release the model checkpoints, as well as some related code and tools, to promote related research. Related literature is reviewed in Appendix B.

\section{Design Overview of FLM-101B}
\label{sec:design}

In this section, we provide an outline of FLM-101B, detailing its architecture, pre-training methods, and configuration specifics.

\subsection{Architecture}

\paratitle{Backbone.}  Among the many existing model architectures, we adopt FreeLM~\cite{freelm} as the backbone for our models, with modifications. FreeLM is based on GPT \cite{radford2019language}, a transformer-like architecture with a decoder-only configuration. Different from GPT, FreeLM features two pre-training objectives: the \textit{language} objective and the \textit{teacher} objective (Section \ref{sec:design:setup}).
We preserve the GPT-3-style transformer block designs without incorporating the later modifications from \llama series. We employ the tokenizer derived from GPT-4, characterized by a vocabulary size of $100,256$.

\paratitle{xPos Integration.} To enhance long sequence modeling, we integrate the Extrapolatable Position Embedding~(xPos) \cite{xpos}. This innovation draws inspiration from RoPE \cite{rope}, which aims to improve the length extrapolation ability by introducing an exponential decay into the rotation matrix.

\paratitle{Model Sizes.} Benefiting from our \textit{growth strategy}, the we produce three models with 16B, 51B, and 101B (\ie FLM-101B) parameters in a single training. The training process is carried out in a progressive manner by growing a 16B model to 51B, and then 101B.

\subsection{Pre-Training Setup}
\label{sec:design:setup}

\paratitle{FLM-101B.}
By design, FLM-101B is an English-Chinese bilingual model. It mixes English and Chinese corpora at a ratio of approximately $53.5\%:46.5\%$. 
Inspired by the finding that instruction data can augment LLMs' comprehension capabilities~\cite{instructgpt}, we integrate multi-task instructionally prompted data: OIG~(Open Instruction Generalist)~\footnote{\url{https://huggingface.co/datasets/laion/OIG}} and COIG~(Chinese Open Instruction Generalist)~\footnote{\url{https://huggingface.co/datasets/BAAI/COIG}}, in the pre-training stage. 

\paratitle{eFLM-16B.}
To analyse the effect of domain-specific knowledge data (Section \ref{sec:knowledge}), we apply the FreeLM teacher signals \cite{freelm} to enhance factual capability. Due to computational cost, we incorporate these signals only in the smallest 16B model. This enhanced model is named eFLM-16B. 

Specifically, we employ two emojis: \includegraphics[scale=0.65]{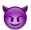} (U+1F621) and \includegraphics[scale=0.65]{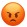}~(U+1F608)~\footnote{\url{https://apps.timwhitlock.info/emoji/tables/unicode}}, from the vocabulary to replace the original binary classification labels \cite{freelm}. For the teacher signal, we supervise only on the emoji tokens, unifying the objective with language modeling. Moreover, we discard the original multi-task alternating approach and completely mix the samples from both sides in every batch. This strategy can enhance the consistency of data sampling distribution as well as improve training stability.

\begin{table}[h]
  \centering
  \caption{\textbf{Partial configurations for different growth stages}.}
  \vspace{1.3ex}
  \scalebox{0.8}{
    \begin{tabular}{c|rrccr}
    \toprule
    \multicolumn{1}{c|}{Params} & \multicolumn{1}{c}{Learning} & \multicolumn{1}{c}{Warmup} & \multicolumn{1}{c}{Batch Tokens} & \multicolumn{1}{c}{Time} & \multicolumn{1}{c}{Tokens} \\
    \multicolumn{1}{c|}{(billion)} & \multicolumn{1}{c}{Rate} & \multicolumn{1}{c}{(samples)} & \multicolumn{1}{c}{(million)} & \multicolumn{1}{c}{(day)} & \multicolumn{1}{c}{(billion)} \\
    \midrule
    16    & $4e-4$  & 4,608,000 & 4.72  & 9.63  & 245.37 \\
    51    & $3.4e-4$ & 230,400 & 4.72  & 5.37  & 39.64 \\
    101   & $2e-4$  & 230,400 & 4.31  & 6.54  & 26.54 \\
    \bottomrule
    \end{tabular}%
}
  \label{tab:config_part}%
\end{table}%

\begin{table*}[t]
  \centering
  \caption{\textbf{Parallel strategies and throughput for different growth stages.} For NVIDIA A800 GPUs, the peak theoretical FLOPs per second is 312 teraFLOPs/sec. Gradient accumulation is applied for the large global batch size.}
  \vspace{0.8ex}
  \scalebox{0.9}{
  \resizebox{\textwidth}{!}{
    \begin{tabular}{c|ccccccc}
    \toprule
    \multicolumn{1}{c|}{Params} & \multicolumn{1}{c}{Tensor} & \multicolumn{1}{c}{Pipeline} & \multicolumn{1}{c}{Data} & \multicolumn{1}{c}{Number} & \multicolumn{1}{c}{Batch} & \multicolumn{1}{c}{teraFLOP/s} & \multicolumn{1}{c}{FLOPs} \\
    \multicolumn{1}{c|}{(billion)} & \multicolumn{1}{c}{Parallel Size} & \multicolumn{1}{c}{Parallel Size} & \multicolumn{1}{c}{Parallel Size} & \multicolumn{1}{c}{of GPUs} & \multicolumn{1}{c}{Size} & \multicolumn{1}{c}{per GPU} & \multicolumn{1}{c}{Utilization} \\
    \midrule
    16    & 2     & 1     & 96    & 192   & 2304  & 162   & 51.90\% \\
    51    & 4     & 2     & 24    & 192   & 2304  & 160   & 51.30\% \\
    101   & 4     & 4     & 12    & 192   & 2160  & 165   & 52.88\% \\
    \bottomrule
    \end{tabular}%
    }
    }
  \label{tab:parallel}%
\end{table*}%

\subsection{Growth Strategy} 

The essence of the low cost in scaling up FLM-101B is the \textit{growth strategy}.  Specifically, we train three models,  with 16B, 51B, and 101B parameters, respectively, in a sequential manner. Each model inherits knowledge from its predecessor.  This is contrary to the common practice that the models of different sizes are trained independently~\cite{llama, llama-2}.

\paratitle{Function-preserving Growth.}
Function preservation means that before and after growth, the models yield consistent outputs given the same arbitrary inputs. This property has proven beneficial for both knowledge inheritance~\cite{net2net, bert2bert, staged} and training stability~\cite{msg}. The growth operators used in FLM-101B training originate from Masked Structural Growth (MSG)~\cite{msg}, with adaptation. Specifically, to adapt these operators to the multi-node 3D parallel framework, we implement them by extending the model structures offline and reloading the checkpoint when the next stage starts.

\paratitle{Schedules and Cost-Effectiveness.}
Model growth scheduling is a trade-off between the pros and cons inherent to models of different sizes \cite{msg}: a smaller model is faster in computation, enabling more rapid consumption of training data for broader commonsense knowledge; conversely, a larger model is better in the reduction of loss per step, indicating a deeper understanding of the nuanced linguistic patterns. Based on the speed test results and total budget, we train the 16B model with 245.37B tokens, the 51B model with 39.64B tokens, and the 101B model with 26.54B tokens. The \textit{billion tokens per day} of different sizes are listed in Table \ref{tab:config_part}. Under this growth schedule, the total time cost for training FLM-101B is 21.54 days, which is 72\% time-saving (or a 3.56x speedup) compared to training a 101B model from scratch (76.74 days estimated). This is consistent with our motivations depicted in Figure~\ref{fig:growth}.

\subsection{The Parallelism Setup and Model Configurations}

\paratitle{The Parallel Strategies.}
FLM-101B is trained on a cluster of 24 DGX-A800 GPU (8×80G) servers. We employ a 3D parallel strategy for optimal throughput, including the standard data parallelism~\cite{parallel-computation}, tensor model parallelism~\cite{DBLP:journals/corr/abs-1909-08053}, and pipeline model parallelism~\cite{DBLP:journals/corr/abs-2104-04473}.
Moreover, by employing sequence parallelism~\cite{korthikanti2022reducing}, we slice the inputs to the Transformer core's LayerNorm and Dropout layers along the sequence length dimension, leading to additional savings in GPU computational resources and memory utilization.
We also utilize the Megetron-LM~\footnote{\url{https://github.com/NVIDIA/Megatron-LM}} implementation of the distributed optimizer~\cite{DBLP:journals/corr/abs-1910-02054} to further reduce GPU memory consumption, which evenly distributes the optimizer states across data parallel ranks.

Table~\ref{tab:parallel} shows the parallelism configurations and training throughput in each stage of FLM-101B training. In different stages, we configure different Tensor Parallel $\times$ Pipeline Parallel sizes to achieve higher efficiency. The single-GPU throughput for all three training stages consistently exceeds 160 teraFLOPs/sec with a utilization rate of at least 51.3\%.
For comparison, GLM-130B achieves 135 teraFLOPs/sec~\cite{glm-130b} with a 42.27\% utilization rate. 
We can also find that FLM-101B has a higher FLOP utilization rate than Megatron-LM~\cite{korthikanti2022reducing} under a similar model size.

\paratitle{FLM-101B Configurations.}
The FLM-101B model is structured with a hidden state dimension of $10,240$, a layer number of 80, a context window of 2,048 tokens, 80 attention heads, and a vocabulary size of $100,256$.
FLM-101B uses the AdamW optimizer~\cite{DBLP:journals/corr/abs-1711-05101} with $\beta_1$ = 0.9 and $\beta_2$ = 0.95. A cosine learning rate schedule is employed, leading to a final learning rate of $6e-6$. We use a weight decay of 0.1 and gradient clipping of 1.0.

Table~\ref{tab:config_part} presents part of the hyperparameters used in different growth stages. In each growth stage, we approximately inherit the previous learning rate and adhere to the same schedule. The learning rate at the beginning of each stage is reported in the table. In the 16B stage, 4,608k samples are used for learning rate warmup, while in later growth stages, we use fewer samples of 230.4k. Note that we do not apply batch size warmup because we address the stability issue in a different manner, detailed in Section \ref{sec:training}. 

\begin{figure}
    \centering
    \includegraphics[scale=0.34]{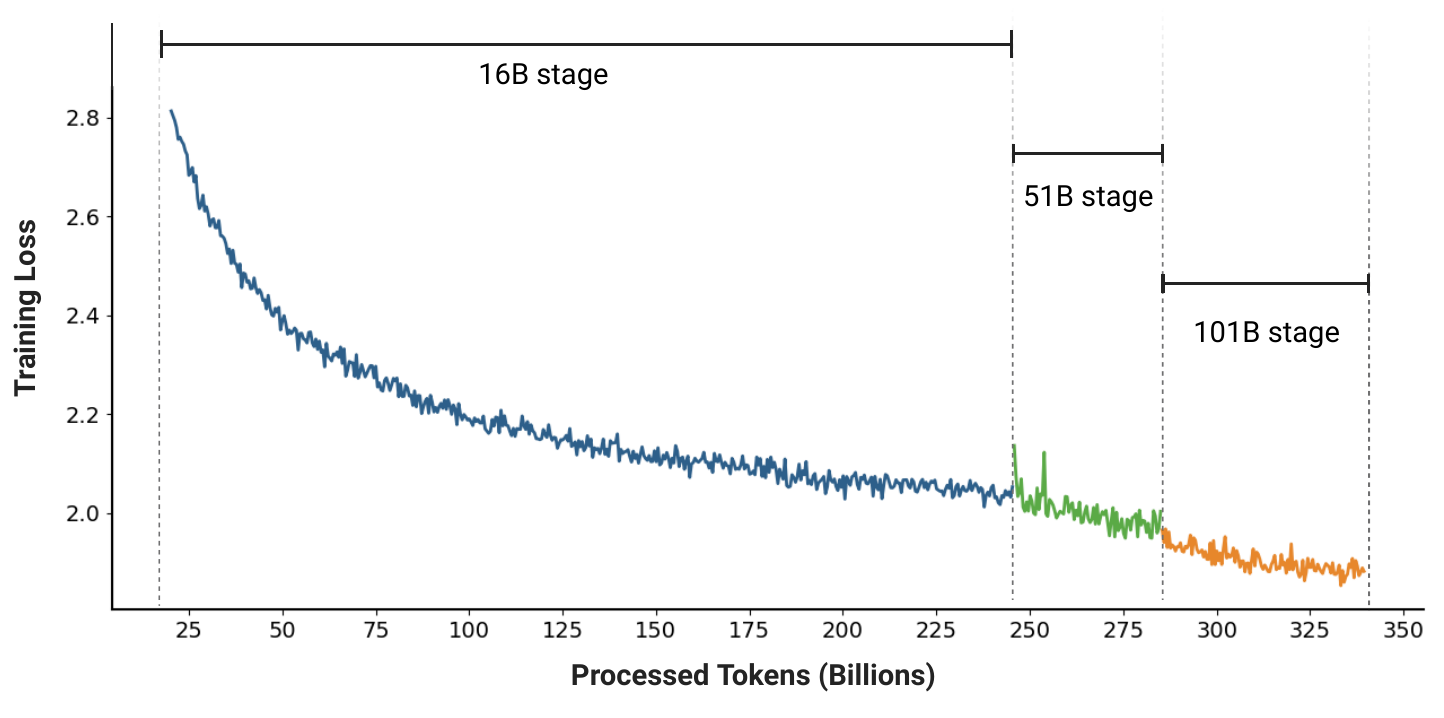}
    \caption{\textbf{Training loss for FLM-101B models.}}
    \label{fig:whole-loss-vs-tokens}
\end{figure}

\begin{table*}[thbp]
\centering
\caption{Carbon emissions of our proposed model, FLM-101B, and other well-known LLMs. For details, please see the corresponding references. The definitions of TDP, net $tCO_2e$, and their formulas are the same as \cite{carbon_compute}.}
\setlength{\tabcolsep}{0.2mm}{
\scalebox{0.95}{
\begin{tabular}{l|cccccc}
\toprule
\footnotesize
Model        & \makecell{GPT-3 \\\cite{gpt3}} & \makecell{Gopher \\\cite{gopher}} & \makecell{PaLM \\\cite{palm2}} & \makecell{GLM-130B \\\cite{glm-130b}} & \makecell{Llama-2 \\\cite{llama-2}} & FLM-101B \\
\midrule
Params & 175B & 280B & 540B & 130B & 70B & 101B \\
GPU Hours & 3.55e6 & 3.77e6 & 8.40e6 & 1.11e6 & 1.72e6 & \textbf{1.01e5} \\
Chip Power/TDP & 330 & 283 & 378.5 & 400 & 400 & 400 \\
Energy (MkWh) & 1171 & 1066 & 3179 & 444 & 688 & \textbf{40} \\
net $tCO_2e$ & 552 & 380 & 271 & 257 &  291 & \textbf{26} \\
\bottomrule
\end{tabular}
}
}
\label{tab:carbon}
\end{table*}
\section{Training Stability of FLM-101B}
\label{sec:training}
Models beyond 100B parameters \cite{bloom, glm-130b} usually suffer from a bunch of notorious stability issues including loss divergence, gradient explosion, and numerical overflow/underflow. This not only inflates the cost of searching for feasible hyperparameters like optimal learning rates, but also intensifies ongoing maintenance during training, such as babysitting, issue resolution, data adjustment, and rebooting. Moreover, this makes the budget of the whole project unpredictable. In this section, we introduce details for mitigating these issues.

\paratitle{Predictable Scaling.} The Tensor Programs theories \cite{TP4,TP4b} unveil the universal relations of the training dynamics with the \textit{model width} tending to infinite. This results in a parameterized mapping for certain classes of hyperparameters between a small model and its larger counterparts, which is termed $\mu$P \cite{TP5}. Two important insights are: (i) \textit{The wider, the better}: theoretically, under $\mu$P transfer, a wider model will always yield lower loss than its narrower counterparts when exposed to identical data \cite{TP5}. As a direct corollary, if a narrow model converges, its wider counterparts will always converge. (ii) \textit{Loss prediction}: the loss value of a large model is predictable using the loss of its smaller counterparts \cite{GPT-4}. $\mu$Scaling \cite{mu-scaling} open-sources a methodology through which loss prediction can be achieved by combining $\mu$P \cite{TP5} and (a modified) scaling law \cite{scaling-law, scaling-law-2, chinchilla}.

Based on these findings, our method to solve training stability is as follows: we first determine the data distribution before the FLM-16B training starts. Next, we perform a grid search on three hyperparameters including the learning rate, initialization standard deviation, and the softmax temperature in the output layer. This grid search is performed with a small \textit{proxy model} (less than $100M$) with a hidden state dimension (``model width'') of 256 and a head number of 2. All the other structural configurations and training data are identical to those of FLM-16B. A single run of grid search takes 24.6 hours with data parallelism on 6 nodes, which is equivalent to 6 hours per run given our 24-node infrastructure. Finally, we find a group of well-performing hyperparameters: learning rate = $4e-4$, standard deviation = $1.6e-2$, and softmax temperature = $2.0$. Transferring these hyperparameters to the 16B model via $\mu$P \cite{TP5} led to a seamless training experience \textit{devoid} of instabilities. Combined with MSG \cite{msg}, we also witness no post-growth divergence in FLM-51B and FLM-101B. 

Our implementations of $\mu$P are largely consistent with those in $\mu$Scaling \cite{mu-scaling}, with modifications to handle the rotary embedding. Thus, the value range of FLM-16B loss is also predictable with the results from multiple proxy widths at the same steps. Mixed-precision training is applied to save run-time memory and reduce time costs. Specifically, we choose \textbf{Bfloat16} instead of FP16 due to its superior precision for values approaching zero, making it more suitable for $\mu$P. As a result, we do not encounter the FP16 underflow issue reported by \cite{TP5}. Moreover, Bfloat16 negates the need for loss scale adjustments, making our training procedure more promising and reproducible. 

The full training loss curve is presented in Figure \ref{fig:whole-loss-vs-tokens}. We observe that the loss curve becomes steeper after each growth. It matches the intuition that a larger model is better in loss reduction per step. The whole training procedure is robust and predictable: even though the 51B stage is short with only 40B tokens, the 101B training remains stable. This supports the effectiveness of the growth strategy.

\begin{table*}[thbp]
\centering
\caption{\textbf{Performance of FLM-101B and baselines including \llama series and GLM-130B.} We list the estimated floating-point operations~($zetta=10^{21}$) of the training process for reference.}
\scalebox{0.95}
{
\begin{tabular}{l|rl|ccccc}
\toprule
Model              & \multicolumn{2}{c|}{Cost~(zettaFLOPs)} & Average & ARC   & HellaSwag & MMLU  & TruthfulQA \\
\midrule
\llama-2~(13B) & 201.37            & ($\pm$28.77)           & 58.66   & 59.39 & 82.13     & 55.77 & 37.38      \\
\llama-2~(7B)  & 106.60            & ($\pm$15.23)           & 54.32   & 53.07 & 78.59     & 46.87 & 38.76      \\
\llama~(13B)   & 94.81             & ($\pm$13.54)           & 56.08   & 56.23 & 80.93     & 47.67 & 39.48      \\
\llama~(7B)    & 49.54             & ($\pm$7.08)            & 49.72   & 51.02 & 77.82     & 35.71 & 34.33      \\
GLM-130B           & 210.80&                 & 48.11   & 42.15 & 67.91     & 42.59 & 39.80      \\
\midrule
FLM-101B           & 28.22&                  & 43.94   & 39.76 & 66.23     & 28.30$^*$ & 41.47      \\
\bottomrule
\multicolumn{8}{l}{\footnotesize{$^*44.50$ for a knowledge-enhanced eFLM-16B (Section \ref{sec:design:setup}, \ref{sec:knowledge}).}}
\end{tabular}
}
\label{tab:openllm}
\end{table*}

\begin{table*}[thbp]
\centering
\caption{\textbf{Performance of eFLM-16B and baselines on C-eval.} In this table, eFLM-16B refers to the professional-knowledge-enhanced FLM-16B. Note that C-Eval leaderboard only keeps one decimal place for the evaluation results.}
\scalebox{0.95}{
\begin{tabular}{l|cccccc}
\toprule
Model        & Average & Average (Hard) & STEM & Social Science & Humanities & Others \\
\midrule
GPT-4        & 68.7 & 54.9        & 67.1 & 77.6           & 64.5       & 67.8   \\
ChatGPT      & 54.4 & 41.4        & 52.9 & 61.8           & 50.9       & 53.6   \\
GLM-130B     & 44.0   & 30.7        & 36.7 & 55.8           & 47.7       & 43.0     \\
\midrule
eFLM-16B      & 46.1 & 28.9        & 38.3 & 53.7           & 46.8       & 52.6   \\ 
\bottomrule
\end{tabular}
}
\label{tab:c-eval}
\end{table*}

\section{Benchmark Evaluation}
\label{sec:pro-exp}

Many existing benchmarks ~(\eg Open LLM\footnote{\url{https://huggingface.co/spaces/HuggingFaceH4/open_llm_leaderboard}.}) focus on assessing the knowledgeability of LLMs. In this section, we discuss the results of FLM on these benchmarks. We believe that knowledge alone might not comprehensively reflect LLM's capability (see Section \ref{sec:knowledge} for more details). Thus, in addition to the common benchmark evaluation, we borrow the concept of IQ tests and evaluate LLMs with some specific cognitive tasks in Section~\ref{sec:iq-exp}.

\paratitle{Cost Estimation Method.}
Due to the considerable computational expense of LLMs, we also emphasize their associated costs in our experimental results. However, it is difficult to directly compare the actual cost of LLMs due to their different infrastructures and prices. 
To objectively compare training costs, we use the FLOPs for training as the cost estimation index, which is estimated from the model's hyperparameters, configuration, and training data~\cite{DBLP:journals/corr/abs-2104-04473}. Since most models do not release the complete training configuration, we estimate FLOPs within a range\footnote{This range originates from the use of checkpoint activation. Please check \cite{DBLP:journals/corr/abs-2104-04473} for more details.}. For monolingual English LLMs, the computational cost of GPT-3 is calculated as $376.41~(\pm53.77)$ zettaFLOPs, and \llama-2 (13B) as $210.37~(\pm28.77)$ zettaFLOPs. 
For bilingual or multilingual models, it is necessary to estimate the cost for each language. 
The total cost of GLM-130B is computed as 421.60 zettaFLOPs. 
As the data ratio of English and Chinese is reported to be 1:1, the cost of GLM-130B for English is $210.80$ zettaFLOPs, and the same for Chinese.
The data ratio of FLM-101B is $53.5\%:46.5\%$ for English and Chinese. The total cost of FLM-101B is computed as $52.76$ zettaFLOPs ($28.22$ zettaFLOPs for English and $24.54$ for Chinese).

\paratitle{Carbon Footprint Analysis.}
An important measurement of a model's environmental impact \cite{greenai} is the carbon footprints originated from the pre-training process. We estimate carbon emission with the methods provided in \cite{carbon_compute}. We summarize the carbon footprint statistics of FLM-101B and well-known LLMs in Table \ref{tab:carbon}. Our model yields only 1/10 pre-training carbon footprint of a typical LLM.

\subsection{Open LLM Evaluation}
\label{sec:openllm-eval}

Open LLM Leaderboard is an open-source project to evaluate the open-sourced LLMs and chatbots. By the time FLM-101B is trained, Open LLM contains four tasks: ARC-Challenge~(ARC for short) \cite{arc}, HellaSwag \cite{hellaswag}, MMLU \cite{mmlu}, and TruthfulQA \cite{truthfulqa}. The Open LLM Leaderboard applies the average score as a metric. All the four tasks require intense knowledge to solve: ARC, HellaSwag, and TruthfulQA depend on commonsense knowledge and Wikipedia, while MMLU contains some questions (\ie STEM) that require domain-specific professional knowledge and intricate reasoning.

Table~\ref{tab:openllm} details the performance of FLM-101B and strong baselines, including \llama series and GLM-130B\footnote{We exclude GPT-3 because it is closed-source. Probability values are unavailable for fair comparison.}. GLM-130B results are achieved by our run on an open-sourced checkpoint.

\paratitle{Results.} On average, FLM-101B achieves a score of $43.94$, reaching over 90\% of the performance of GLM-130B, which has 7 times more FLOPs. Both model underperform the \llama series, potentially due to the first-generation model architectures and less well-refined training data. Note that the FLOPs of FLM-101B is even lower than a 7B \llama model. Going deeper into the nature of these tasks, we further have the following observations: 

(i) MMLU typically requires domain knowledge to solve. In our training, no English textbook or exam data is intentionally used. Nevertheless, our eFLM-16B (Section \ref{sec:design:setup}) variant, incorporated with these knowledge via FreeLM objectives, outperforms GLM-130B with only 16B parameters.

(ii) As aforementioned, TruthfulQA, ARC, and HellaSwag emphasize more on common sense and Wiki-level knowledge; their performances improve with the increased amount of data and the reduction of training loss. With less than 0.16T English data (about 1/10 of \llama-2), FLM-101B already achieves the best accuracy of $41.47$ among all the baselines on TruthfulQA. On ARC and HellaSwag, FLM-101B is comparable to GLM-130B with a similar amount of English data (approximately 0.2T). Also, the training data of GLM-130B includes ARC and HellaSwag, as expressly claimed in \cite{glm-130b}. In our understanding, for FLM-101B, improvement can be expected on these three tasks if exposed to more training data.

\begin{table*}[thbp]
  \centering
  \caption{\textbf{Performance of the three stages of FLM on Open LLM.} To reduce the computational cost during evaluation, we sample $20\%$ and $30\%$ items for HellaSwag and MMLU tasks, respectively.}
  \scalebox{1.0}{
    \begin{tabular}{cccccccc}
    \toprule
    Parameters & Training Data & \multicolumn{1}{l}{Average} & \multicolumn{1}{l}{ARC} & \multicolumn{1}{l}{Hellaswag} & \multicolumn{1}{l}{MMLU} & \multicolumn{1}{l}{TruthfulQA} \\
    \midrule
    16B   & 245.37B  & 39.19 & 32.25 & 58.57 & 27.02 & 38.92 \\
    51B   & 39.64B   & 41.79 & 35.32 & 64.04 & 27.66 & 40.12 \\
    101B  & 26.54B   & 44.41 & 39.76 & 67.88 & 28.54 & 41.47 \\
    \bottomrule
    \end{tabular}
    }
  \label{tab:parameter}
\end{table*}

\subsection{Evaluation on the Professional Knowledge-Enhanced Model}
\label{sec:knowledge}

We conduct experiments on a knowledge-enhanced version~(eFLM-16B, detailed in Section \ref{sec:design:setup}) of the FLM to validate the effect of domain-specific knowledge data on benchmark results. We continue to train the smallest FLM-16B with teacher signals \cite{freelm} from a combination of (i) part of the auxiliary training data of MMLU \cite{mmlu}, (ii) exam questions in similar domains and formats to C-Eval \cite{c-eval}~\footnote{C-Eval can be considered as a Chinese version of MMLU.}, and (iii) other domain knowledge data. 
Note that eFLM-16B is not a typical fine-tuning with instruct data which may affect the language modeling capability of LLM. We preserve both language and teacher signals with the corresponding data in this continue-training. The MMLU result is in the footnote of Table \ref{tab:openllm}. Table~\ref{tab:c-eval} lists the result of eFLM-16B and baselines on C-Eval.

\paratitle{Results.}
Enhanced with professional knowledge, significant improvements are observed.
On MMLU tasks, the incorporation of professional knowledge data results in a score of $44.50$ for eFLM-16B~(see Table~\ref{tab:openllm}), which surpasses GLM-130B ($42.59$), a model that also incorporated multi-task data in the related domain \cite{glm-130b}. For comparison, the MMLU score is $27.02$ for the un-enhanced FLM-16B. On C-Eval tasks~\footnote{The scores are achieved on the test set by submitting to the C-Eval platform.}, we observe that eFLM-16B performs better than GLM-130B by about 2 points. For comparison, the average C-Eval score of the vanilla FLM-16B is $27.0$, which underperforms GLM-130B.
 These results suggest that evaluation with professional knowledge may not fully reflect the capability of LLMs, particularly when different LLMs are trained with different data collections, and some may not come with a clear list.

\subsection{Evaluation of the Growth Strategy}

Our core method for reducing computational cost is the growth strategy. We would like to answer the questions of whether our growth strategy is effective in knowledge inheritance, and how model capabilities grow with size. We evaluate the performance of FLM on all the stages: 16B, 51B, and 101B.
Table~\ref{tab:parameter} shows the performance of FLM models at each stage.

\paratitle{Results.} 
As expected, the performance of FLM improves with the increase in model size. FLM-101B achieves the best performance on almost all tasks. This means that our model inherits knowledge from the previous stage after each growth. We also observe that the 101B model improves the performance scores more significantly than the 51B model, with less data. This indicates that the models are successfully incorporating new weights in training after growth, and taking advantage of larger parameter counts. The performance on ARC and HellaSwag increases steadily and significantly, which corresponds well to the decline of the model loss. Again, as we expected in Section \ref{sec:openllm-eval}, as more training data is processed, FLM’s performance on Open LLM improves.

\section{Evaluation Inspired by IQ Tests}
\label{sec:iq-exp}

Section~\ref{sec:pro-exp} presents the evaluation of existing benchmarks, focusing on knowledge. As we discussed in Section~\ref{sec:intro} and \ref{sec:knowledge}, knowledge could not fully reflect the Intelligence Quotient (IQ) of LLMs. As a supplement, we conduct a series of IQ-test task evaluation in this section. For IQ evaluation, we make necessary modifications to existing datasets \cite{DBLP:journals/corr/abs-2305-08298, babi, big-bench} or generate new synthetic datasets. Specifically, the IQ test mainly considers four aspects: \textit{symbolic mapping}, \textit{rule understanding}, \textit{pattern mining}, and \textit{anti-interference}. A common key property of these tasks is that they are dependent on the inference and generalization in a new context, instead of the previously-learned knowledge.

\paratitle{Compared Methods.}
Borrowing psychological ideas that the measurement of IQ is dependent on age \footnote{\url{https://ocw.mit.edu/ans7870/9/9.00SC/MIT9_00SCF11_text.pdf}, page 367.}, we mainly consider existing models trained with similar amounts of data to FLM-101B. As a milestone of LLM development, GPT-3 (175B) \cite{gpt3} proposed in-context learning for the first time. GLM-130B \cite{glm-130b} is the first open English-Chinese bilingual LLM. Hence, we select them as baseline models. Both models are trained with 300~\textasciitilde 400 billion tokens, which are in the same range as ours. GPT-3 focuses on English, and is not included in the Chinese-related evaluation~(\ie CLUE-IQ). The results of GPT-3 are achieved by API. GLM-130B is evaluated with its open-sourced checkpoint.

\paratitle{Tasks, Data, and Results.}
We curate evaluation benchmarks regarding four cognitive capabilities. We summarize the data and results in this section. For details in task definition and data collection process, please see Appendix A.1 to A.4.

\textit{Symbolic Mapping.} 
An existing study~\cite{DBLP:journals/corr/abs-2305-08298} points out that textual classification tasks~(\eg sentiment classification) often lack generalization. Considering this, we use a symbolic mapping method to replace the original categorical labels with symbols that are unlikely to be seen in any training data. Hence, we can evaluate the LLMs' language understanding ability as well as the generalization abilities to a new context. We form our evaluation task as in-context learning with few-shot examples for each label. We construct two symbolic mapping datasets, namely SuperGLUE-IQ and CLUE-IQ, built on SuperGLUE \cite{superglue} and CLUE \cite{clue}, respectively. Examples are illustrated in the Appendix (Figure 1).

Results on SuperGLUE-IQ and CLUE-IQ are presented in Table~\ref{tab:superglue} and Table~\ref{tab:clue}, respectively. With less computation by one magnitude, FLM-101B achieves comparable performance with GPT-3 on SuperGLUE-IQ and outperforms GLM-130B on CLUE-IQ.

\begin{table}[h]
\centering
\caption{\textbf{Performance on SuperGLUE-IQ of GPT-3, GLM-130B, and FLM-101B.} Cost is computed in zettaFLOPs.}
\scalebox{0.73}{
\begin{tabular}{l|c|ccccc}
\toprule
Model    & \makecell{Cost} & Average  & BoolQ & WiC   & RTE & WSC   \\ \midrule
GPT-3    & 376.41~($\pm$53.77) & 47.60 & 50.84 & 53.33  & 48.38 & 37.86   \\
GLM-130B & 210.80 & 48.19 & 40.13 & 48.67 & 47.65 & 56.31 \\
FLM-101B & 28.22 & 46.76 & 49.50  & 50.33 & 48.38 & 38.83 \\ \bottomrule
\end{tabular}
}
\label{tab:superglue}
\end{table}

\begin{table}[h]
\centering
\caption{\textbf{Performance on CLUE-IQ for GLM-130B and FLM-101B.} Cost is computed in zettaFLOPs.}
\scalebox{0.77}{
\begin{tabular}{l|c|ccccc}
\toprule
Model   & \makecell{Cost} & Average & AFQMC & CSL   & OCNLI & \makecell{CLUE\\WSC\\2020} \\ \midrule
GLM-130B & 210.80 & 39.96 & 33.33 & 53.85 & 34.0  & 38.67       \\
FLM-101B & 24.54 & 42.07 & 38.33 & 55.29 & 27.33 & 47.33       \\ \bottomrule
\end{tabular}
\label{tab:clue}
}
\end{table}

\textit{Rule Understanding.}
We consider the understanding and execution of rules being a strong indication of reasoning capability. To this end, we design rule understanding evaluation. Note that this test is different from reasoning based on the chain of thought. Detailed discussion is provided in Appendix A.2.

We curate data for two subtasks: \textit{Counting} (0-shot) and \textit{String replacement} (4-shots). 

\begin{table}[thbp]
\centering
\caption{ \textbf{Performance of FLM-101B, GPT-3, and GLM-130B on rule understanding tasks.} }
\label{tab:iq:rule}
\scalebox{0.8}{
\begin{tabular}{l|c|ccc}
\toprule
Model    & Average & Counting & \makecell{Replace\\Lowercase} & \makecell{Replace\\Word} \\ \midrule
GPT-3    & 86.03 &82.43    & 80.67                          & 95.00                        \\
GLM-130B & 71.49 &60.81    & 69.67                          & 84.00                     \\
FLM-101B & 76.42 &69.59    & 64.00                          & 95.67                     \\ \bottomrule
\end{tabular}
}
\end{table}

Table~\ref{tab:iq:rule} shows the performance of our proposed FLM-101B against GPT-3 and GLM-130B on rule understanding tasks. For Counting, FLM-101B achieves 69.59\%, about 9 points better than GLM-130B. 
GPT-3 wins the first place in counting and Replace-Lowercase, and second place in Replace-Word. This is potentially because GPT-3 is the largest model.

\textit{Pattern Mining.}
Pattern Mining evaluation is common in IQ tests. In detail, it is the induction and deduction of the patterns emerging in a new context. 

We build a benchmark with three tasks~(\ie Head \& Tail, Full Repeating, and Head Slicing) for evaluation. Figure 2 in the Appendix shows examples of these tasks. Each task is 5-shot and contains 100 instances. 

\begin{table}[thbp]
\centering
\caption{ \textbf{Performance of FLM-101B, GPT-3, and GLM-130B on pattern mining tasks.} }
\scalebox{0.8}{
\begin{tabular}{@{}lcccc@{}}
\toprule
Model   & Average & Head \& Tail & Full Repeating & \multicolumn{1}{r}{Head Slicing} \\ \midrule
GPT-3   & 70.00 & 61.00           & 92.00     & 57.00                          \\
GLM-130B & 53.00 & 38.00           & 70.00     & 51.00                          \\
FLM-101B & 64.67 & 52.00           & 79.00     & 63.00                          \\ \bottomrule
\end{tabular}
}
\label{tab:iq:pattern}
\end{table}

Table~\ref{tab:iq:pattern} lists the experimental results of our FLM-101B against the baselines on pattern mining tasks. On all three tasks, FLM-101B outperforms GLM-130B by a large margin. For the Head \& Tail and Full Repeating tasks, FLM-101B is a few points behind GPT-3, but outperforms the latter on the Head Slicing task. Considering the computational cost, FLM-101B exhibits noticeable abilities.

\textit{Anti-interference.}
Anti-interference capability is critical for finding and utilizing information that is truly related to a specific goal, in an unseen and noisy context (Appendix Figure 3). As suggested by the cocktail party problem in speech recognition~\cite{DBLP:journals/jzusc/QianWCWY18}, we consider anti-interference ability to be important for intelligent agents. We conduct anti-interference evaluation in three task types: Multiple Key Retrieval, Single Supporting Fact Tracking, and Two Supporting Facts Tracking, as exemplified in Figure 3 in the Appendix.

\begin{table}[thbp]
\centering
\caption{ \textbf{Performance of FLM-101B, GPT-3, and GLM-130B on anti-interference evaluation.}}
\vspace{0.6ex}
\scalebox{0.47}{
\resizebox{\textwidth}{!}{
    \begin{tabular}{lcccc}
    \toprule
    Model  & Average & \makecell{Multiple\\Key\\Retrieval} & \makecell{Single\\Supporting\\Fact} & \makecell{Two\\Supporting\\Facts} \\ \midrule
    GPT-3   & 70.11 & 92.67                   & 78.33                  & 39.33                \\
    GLM-130B & 53.56 & 77.67                   & 56.33                  & 26.67                \\
    FLM-101B & 60.11 & 89.00                      & 59.00                     & 32.33                \\ \bottomrule
    \end{tabular}
}
}
\label{tab:iq:anti}
\end{table}

Table~\ref{tab:iq:anti} shows the evaluation results on anti-interference. FLM-101B achieves the second-best passing rates with an advantage of around 7\% compared to GLM-130B.

\paratitle{IQ Test Conclusion.} 
On our four additional evaluations inspired by the IQ tests, FLM-101B outperforms GLM-130B and obtains competitive results compared to GPT-3 in some tasks with much lower costs. Except for the impacts of training data, the superiority may be owed to a story that in the growth strategy, the smaller models in early stages refine a more efficient searching space, which keeps taking effect when the model grows larger with increased generalization ability.

\section{Conclusions, Limitations, and Future Work}
\label{sec:con}

In this paper, we introduce FLM-101B, an open-sourced LLM that is successfully trained from scratch within a \$100,000 budget. The key idea of reducing the training cost of FLM-101B is to break through the fixed number of model parameters via a growth strategy. Experimental results on knowledeg-oriented and IQ-related benchmarks show that FLM-101B is comparable to strong baseline models with less computational cost. Note that harmful contents may be induced from the open-sourced checkpoint, which do not represent the opinions of the authors.

Due to resource issues, the limitations of our work include inadequate exploration and comparison for different growth schedules, growth operators, and amount of data. For future work, we believe that our exploration on the growth strategy as well as training stability would potentially be beneficial for future attempts of further scaling up LLMs, \eg beyond 1T parameters.

\bibliography{custom}

\end{document}